\documentclass[lettersize,journal]{IEEEtran}       
\usepackage{amsmath,amsfonts}
\usepackage{bm}
\usepackage{algorithmic}
\usepackage{algorithm}
\usepackage{array}
\usepackage[caption=false,font=normalsize,labelfont=sf,textfont=sf]{subfig}

\usepackage{textcomp}
\usepackage{stfloats}
\usepackage{url}
\usepackage{verbatim}
\usepackage{graphicx}
\usepackage{cite}
\usepackage{booktabs}
\usepackage{multirow}

\begin{document}

\title{Lean Learning Beyond Clouds: Efficient Discrepancy-Conditioned Optical-SAR Fusion for Semantic Segmentation}

\author{Chenxing Meng$^\dag$, Wuzhou Quan$^\dag$, Yingjie Cai, Liqun Cao, Liyan Zhang, Mingqiang Wei,~\IEEEmembership{Senior Member,~IEEE}
        
\thanks{C. Meng, W. Quan, Y. Cai, L. Cao, L. Zhang, M. Wei are with the School of Computer Science and Technology, Nanjing University of Aeronautics and Astronautics, Nanjing, China,  (e-mail: a1757845799@gmail.com; q.wuzhou@gmail.com; cyjeven@gmail.com; liquncao95@gmail.com; zhangliyan@nuaa.edu.cn; mingqiang.wei@gmail.com).}
\thanks{$^\dag$ These authors contributed equally to this work.}
\thanks{Corresponding author: Mingqiang Wei.}
}

\markboth{IEEE TGRS, VOL. X, NO. X, MARCH 2026}%
{Shell \MakeLowercase{\textit{et al.}}: A Sample Article Using IEEEtran.cls for IEEE Journals}



\maketitle

\begin{abstract}

Cloud occlusion severely degrades the semantic integrity of optical remote sensing imagery. While incorporating Synthetic Aperture Radar (SAR) provides complementary observations, achieving efficient global modeling and reliable cross-modal fusion under cloud interference remains challenging.
Existing methods rely on dense global attention to capture long-range dependencies, yet such aggregation indiscriminately propagates cloud-induced noise.
Improving robustness typically entails enlarging model capacity, which further increases computational overhead.
Given the large-scale and high-resolution nature of remote sensing applications, such computational demands hinder practical deployment, leading to an efficiency-reliability trade-off.
To address this dilemma, we propose EDC, an efficiency-oriented and discrepancy-conditioned optical-SAR semantic segmentation framework. 
A tri-stream encoder with Carrier Tokens enables compact global context modeling with reduced complexity.
To prevent noise contamination, we introduce a Discrepancy-Conditioned Hybrid Fusion (DCHF) mechanism that selectively suppresses unreliable regions during global aggregation.
In addition, an auxiliary cloud removal branch with teacher-guided distillation enhances semantic consistency under occlusion.
Extensive experiments demonstrate that EDC achieves superior accuracy and efficiency, improving mIoU by 0.56\% and 0.88\% on M3M-CR and WHU-OPT-SAR, respectively, while reducing the number of parameters by 46.7\% and accelerating inference by 1.98$\times$.
Our implementation is available at \url{https://github.com/mengcx0209/EDC}.


\end{abstract}


\section{Introduction}
\IEEEPARstart{S}{emantic} segmentation of optical remote sensing (RS) imagery is a cornerstone of Earth observation, enabling pixel-wise understanding of land-cover semantics\cite{li2024rsseg_review}. 
While traditional models such as support vector machines and conditional random fields once dominated the field, they have largely been superseded by deep learning (DL) methods, which have achieved unprecedented performance in complex scenarios\cite{10916803,liu2024optsarreview}.
However, DL-based models remain challenged by cloud occlusion, which invalidates optical observations over large regions\cite{MERANER2020333,ebel2021sen12mscr}.
Persistent cloud cover, especially in tropical and coastal regions, leads to severe spectral and  spatial information loss, hindering timely and accurate land cover monitoring\cite{zhou2024allclear,floresanderson2023cloudfree, LIU2025104567}.

\begin{figure}[t]
  \centering
  \includegraphics[width=1\columnwidth]{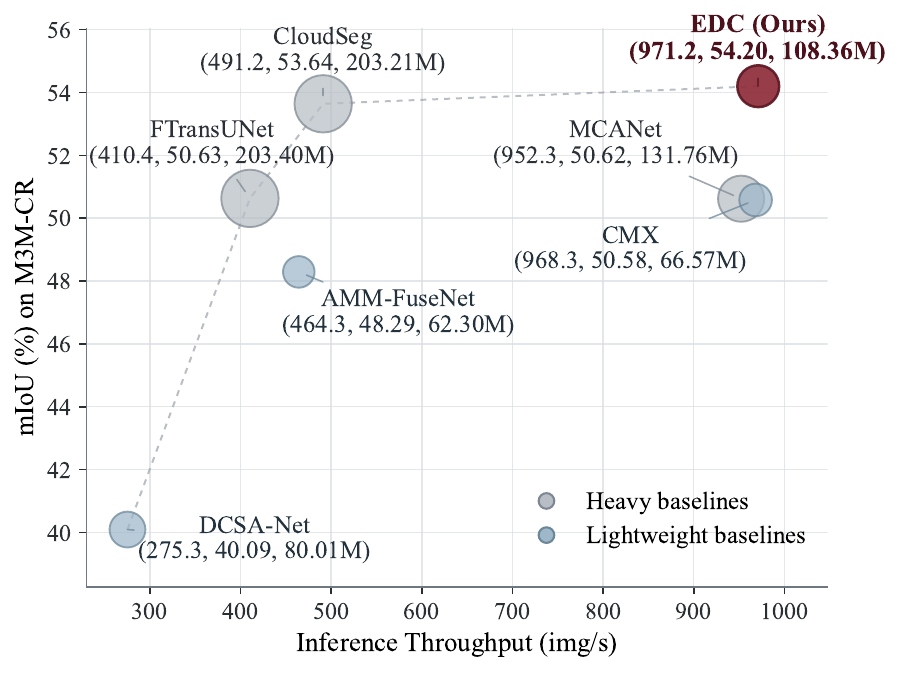} 
  \caption{Efficiency–accuracy comparison on M3M-CR. Bubble size indicates model parameter scale. The annotation beside each bubble gives the triplet (throughput, mIoU, parameters). EDC lies on the favorable Pareto frontier, achieving the best mIoU and the highest inference throughput simultaneously.} 
  \label{fig:pareto}
\end{figure}

Cloud occlusion introduces two related challenges: the loss of semantic context and unreliable local features.
Remote sensing scenes exhibit strong spatial continuity, where land-cover categories (e.g., agricultural parcels, river networks, and urban blocks) extend over large areas and follow structured spatial patterns. 
As a result, reliable interpretation often depends on contextual information beyond the local region.
When large-scale cloud occlusion occurs, optical cues within the affected areas become corrupted, making local observations insufficient for accurate prediction.
Thus, accurate prediction requires long-range contextual dependencies that leverage information from distant clear-sky regions.
However, modeling such long-range context is typically computationally expensive, creating a practical challenge for efficient deployment~\cite{liu2023efficientvit,liu2024vmamba}. As illustrated in Fig.~\ref{fig:pareto}, heavy baselines improve accuracy at the cost of substantially lower inference throughput, whereas lightweight alternatives retain high speed but often sacrifice segmentation quality. 

This tension is particularly evident in global context modeling.
Dense self-attention provides a natural mechanism to capture global dependencies, but its quadratic complexity quickly becomes prohibitive as image resolution increases\cite{wu2021cvt}.
Even with sequence reduction strategies (e.g., reducing the length of K/V with a spatial reduction ratio $R$\cite{xie2021segformer,rs17142508}), the attention cost still grows rapidly for high-resolution dense prediction and large receptive fields.
To enable robust global reasoning, recent methods such as CloudSeg\cite{xu2024_cloudseg} and FTransUNet\cite{ma2024ftransunet} rely on heavy backbones, typically large-scale hierarchical Transformers or hybrid architectures.
As shown in Tab.~\ref{tab:efficiency_miou}, these models often contain hundreds of millions of parameters (e.g., CloudSeg at $\sim$203M) and require substantial FLOPs per patch, which severely limits inference throughput.
In contrast, lightweight architectures usually lack sufficient receptive fields to infer consistent labels under large cloud masses, resulting in fragmented predictions in heavily occluded regions\cite{zhang2023cmxcrossmodalfusionrgbx,li2022_mcanet}.

The second equally critical limitation lies in the reliability of feature fusion.
Cloud occlusion induces spatially varying signal-to-noise ratios: clear-sky regions provide high-fidelity spectral cues, whereas cloud-covered regions are dominated by noise.
However, conventional fusion paradigms often adopt indiscriminate aggregation, treating all spatial locations as equally informative and failing to differentiate between clear and occluded regions.
As a result, cloud-corrupted optical artifacts are fused with stable SAR cues, contaminating the cross-modal representation and reducing the benefit that SAR could otherwise provide in occluded areas.

To address these challenges, we propose \textbf{EDC}, an efficiency-oriented and discrepancy-conditioned progressive spatial-spectral fusion framework that balances segmentation accuracy and computational efficiency. 
First, on the efficiency front, we introduce the \textit{Efficiency-Oriented Multi-Scale Encoder} (EOME), a carrier-token-based tri-stream encoder that learns optical, SAR, and cross-modal representations. 
By routing global context through compact carrier tokens rather than dense token-to-token attention, EOME achieves efficient long-range modeling for inferring semantics in cloud-occluded regions.
Second, we propose a \textit{Discrepancy-Conditioned Hybrid Fusion} (DCHF) mechanism for reliable optical–SAR fusion under cloud occlusion. DCHF leverages pixel-wise cross-modal discrepancy as a reliability cue to suppress cloud-contaminated optical responses and stabilize both spatial interaction and channel recalibration during fusion.
Finally, we adopt a Dual-Task Learning strategy coupled with teacher-guided distillation\cite{kendall2018multitask, hinton2015distilling,rs17223721}. 
An auxiliary cloud removal head forces the network to learn cloud-invariant surface representations, while feature distillation transfers clear-sky semantic patterns from a cloud-free teacher to the student on reliable clear pixels, improving representation alignment and stabilizing semantic learning under cloud interference.
Extensive experiments demonstrate that EDC improves segmentation accuracy while achieving $1.98\times$ higher inference throughput than representative heavy baselines. 
As illustrated in Fig.~\ref{fig:pareto}, EDC lies on a favorable efficiency–accuracy Pareto frontier, achieving the highest mIoU while also delivering the highest inference throughput on M3M-CR dataset with a moderate parameter budget.


In summary, our main contributions are:

1) We propose EDC, a cloud-resilient segmentation framework that balances accurate global context modeling with computational efficiency.
The framework integrates an efficiency-oriented encoder, discrepancy-conditioned fusion, and a dual-task distillation strategy.

2) We design the Efficiency-Oriented Multi-Scale Encoder (EOME), a carrier-token-based tri-stream encoder that captures large-area semantic context with significantly reduced computational cost.

3) We introduce the Discrepancy-Conditioned Hybrid Fusion (DCHF) module, a discrepancy-conditioned gating mechanism that suppresses cloud-contaminated optical responses and stabilizes cross-modal fusion.



\section{Related Work}

\subsection{Efficient Semantic Segmentation in Remote Sensing}
Semantic segmentation serves as a fundamental task in RS interpretation\cite{zhu2017dlrs}. Early approaches were predominantly built upon Convolutional Neural Networks (CNNs). Seminal architectures, such as FCN \cite{long2015fcn}, U-Net\cite{ronneberger2015unet}, and DeepLab series\cite{chen2018deeplabv3plus}, have demonstrated remarkable success in extracting hierarchical features. However, information interaction in CNNs is implicitly propagated through hierarchical stacking. This mechanism hinders the direct modeling of long-range dependencies, resulting in an effective receptive field (ERF) that is often localized and fails to capture the holistic context of large-scale geospatial objects\cite{luo2016erf}.
To overcome the locality of convolutions, Vision Transformers (ViTs)\cite{dosovitskiy2021vit} model global context via self-attention, but the quadratic complexity with respect to token length makes them expensive for high-resolution RS imagery.
Although efficiency-oriented variants\cite{xie2021segformer,FAN2026106848} utilize a sequence reduction strategy to shrink the spatial dimension of key and value sequences, thereby reducing the computational overhead, the cost of global interaction remains a major bottleneck as resolution grows, limiting throughput in dense prediction\cite{Wang31122025}.
This bottleneck is further amplified in multimodal fusion, where multi-stream encoders substantially increase computation\cite{FAN2026106848}.

This dilemma has spurred research into hybrid architectures that seek a better trade-off between efficiency and global modeling \cite{guo2022segnext,liu2024ViTHA}.
Recent hybrid designs\cite{hatamizadeh2024fastervitfastvisiontransformers} decompose global modeling into window-based local attention and a small set of carrier tokens for cross-window exchange. This design decouples local feature extraction from long-range dependency modeling, enabling efficient global context aggregation without dense full-resolution attention. Motivated by the efficiency requirement of multi-stream processing, we adopt a carrier-token-based tri-stream encoder for cloud-resilient representation learning.


\subsection{Multimodal Fusion and Reliability Modeling for Cloud Resilience}

Cloud occlusion poses a persistent challenge in optical remote sensing. To reconstruct the missing land cover semantics beneath clouds, fusing optical imagery with all-weather SAR data has become a standard paradigm for cloud-resilient mapping\cite{MERANER2020333,LI2026103671,schmitt2019sen12ms,ebel2022sen12mscrts}.
Early attempts typically employed early fusion strategies, such as channel concatenation or element-wise summation, to combine features from both modalities\cite{li2022_mcanet,HONG2025104827,kang2022cfnet}. While straightforward, these methods treat all spatial locations equally, failing to account for the varying information quality caused by cloud artifacts. To mitigate the impact of occlusion, state-of-the-art frameworks like CloudSeg\cite{xu2024_cloudseg} have advanced towards multi-task learning. Specifically, CloudSeg incorporates an auxiliary cloud removal task and utilizes knowledge distillation to guide the network in recovering cloud-free semantics\cite{LI2025102649}. 
However, the fusion operation is often still indiscriminate, implicitly treating restored optical features as reliable across all regions. Under thick clouds, restoration can be ill-posed, and corrupted optical cues may propagate through fusion and pollute the cross-modal representation.

These limitations call for reliability-aware fusion that explicitly distinguishes reliable and unreliable regions under spatially varying cloud corruption.
We therefore propose Discrepancy-Conditioned Hybrid Fusion (DCHF), which uses cross-modal discrepancy as a reliability cue to gate information flow based on physical consistency.

\section{Method}
\subsection{Overview}
\begin{figure*}[t]
  \centering
  \includegraphics[width=\linewidth]{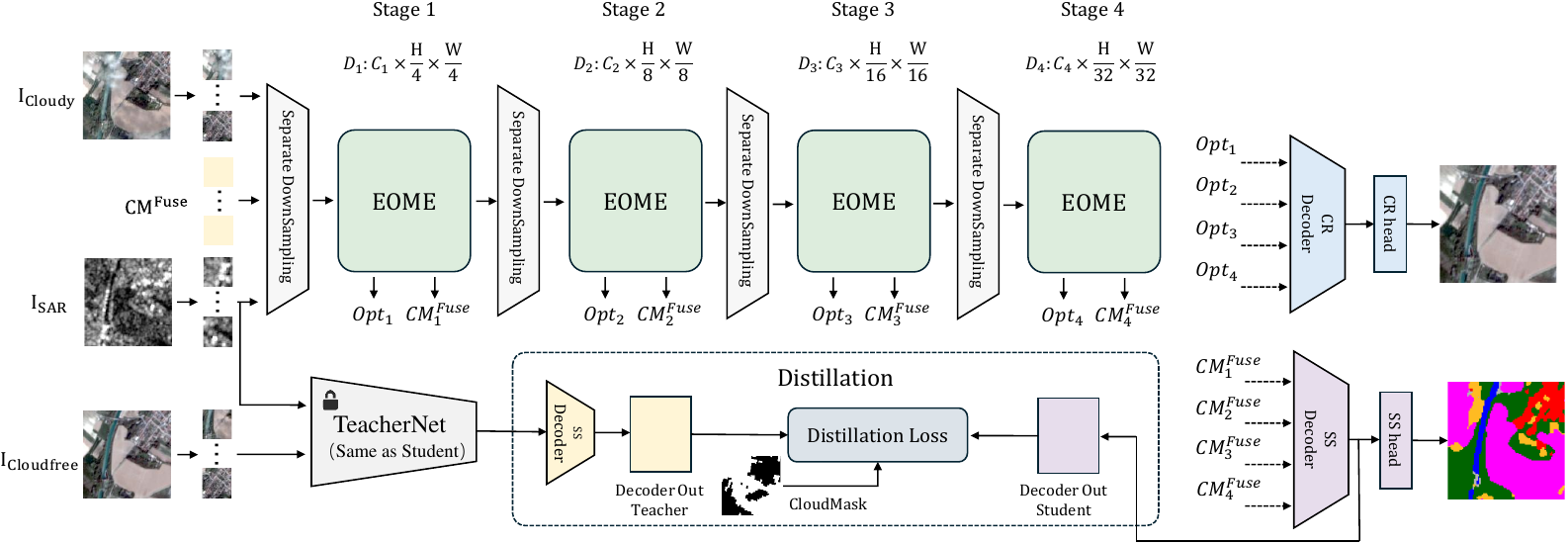}
  \caption{Overall framework of the proposed EDC. The StudentNet takes the SAR image $I_{\text{SAR}}$ and the cloudy optical image $I_{\text{Cloudy}}$ as input. At each stage $i$, the encoder produces an optical-corrected feature $Opt_i$ and a fused semantic feature $CM_i^{\text{Fuse}}$ at the corresponding resolution. Note that $CM_1^{\text{Fuse}}$ is initialized as an empty input (i.e., no prior fused feature is available at the first stage). The cloud-removal (CR) decoder aggregates $\{Opt_i\}_{i=1}^{4}$ to reconstruct the cloud-removed optical image, while the semantic-segmentation (SS) decoder aggregates $\{CM_i^{\text{Fuse}}\}_{i=1}^{4}$ to predict the land-cover map. During training, a TeacherNet with the same architecture is fed with $I_{\text{SAR}}$ and the cloud-free optical image $I_{\text{Cloudfree}}$ and provides supervision to the StudentNet via a cloud-mask-guided distillation loss.}
  \label{fig:overview}
\end{figure*}

As illustrated in Fig.~\ref{fig:overview}, the proposed EDC is formulated as a progressive spatial-spectral fusion framework. To balance long-range context modeling and efficiency under cloud occlusion, we design a lightweight tri-branch student architecture with teacher-guided distillation. The workflow proceeds in three stages:

1) Efficiency-Oriented Multi-Scale Encoding: Instead of relying on dense self-attention with quadratic complexity\cite{vaswani2017attention}, we construct an Efficiency-Oriented Multi-Scale Encoder with three parallel streams (Optical, SAR, and Cross-Modal). With modality-adaptive stems for heterogeneous inputs, the encoder leverages a hierarchical carrier-token mechanism. This design decouples local extraction from global dependency modeling, enabling efficient long-range context aggregation with near-linear complexity.

2) Discrepancy-Conditioned Hybrid Fusion Module: To handle spatially varying feature reliability under cloud occlusion, hierarchical features are fused by DCHF. DCHF computes a discrepancy attention map $\mathbf{A}$ from cross-modal feature differences and uses it to guide both (i) discrepancy-focused spatial interaction and (ii) discrepancy-conditioned global descriptor construction via WGAP. This prevents cloud-contaminated responses from biasing channel statistics. Meanwhile, it emphasizes stable SAR cues in regions with high cross-modal inconsistency before feeding the refined representation into the cross-modal stream.

3) Dual-Task Decoding and Optimization: The refined features are upsampled by a U-Net-style decoder to recover fine-grained boundaries. We train the network with dual-task objectives: a semantic segmentation head predicts the land-cover map, while an auxiliary cloud removal head reconstructs the cloud-free optical image. The auxiliary task regularizes the encoder toward cloud-invariant surface details.

\begin{figure}[t]
  \centering
  \includegraphics[width=0.9\columnwidth]{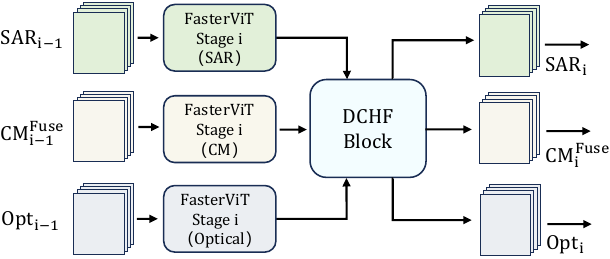}
  \caption{Framework of the proposed Efficiency-Oriented Multi-Scale Encoder.} 
  \label{fig:EOME}
\end{figure}

\subsection{Efficiency-Oriented Multi-Scale Encoder}

Long-range context modeling is essential for cloud-occluded dense prediction, yet a direct implementation via dense self-attention requires full token-to-token interaction. Let $S = H \times W$ denote the number of spatial tokens. Standard multi-head self-attention incurs $\mathcal{O}(S^{2})$ interactions in token length, which becomes prohibitive as resolution increases. SegFormer alleviates this cost by sequence reduction in key/value tokens with a spatial reduction ratio $R$ (i.e., reducing K/V length to approximately $S/R^{2}$), leading to an attention cost on the order of $\mathcal{O}(S^{2}/R^{2})$; however, the computation still grows rapidly for high-resolution dense prediction and large receptive fields\cite{xie2021segformer}.

To reconcile long-range context modeling with computational efficiency, EDC adopts an \emph{efficiency-oriented multi-scale encoder} instantiated by the Hierarchical Attention Transformer (HAT) backbone in FasterViT, and extends it into a \emph{tri-stream} design: an Optical stream, a SAR stream, and a Cross-Modal stream, which is illustrated in Fig. \ref{fig:EOME}. HAT introduces \emph{carrier tokens} to enable cross-window communication at reduced cost: local window tokens are aggregated within windows, while carrier tokens mediate global message passing across windows, thereby preserving long-range context without full quadratic attention over all spatial tokens.

\textbf{Modality-adaptive stem initialization.}
To accommodate heterogeneous inputs while reusing ImageNet pretraining, we adopt a modality-adaptive stem initialization strategy. Given the pretrained stem kernel $\mathbf{W}\in\mathbb{R}^{C_{\mathrm{out}}\times 3\times k\times k}$, we adapt the input channels as follows: for 4-band optical input (RGB+NIR), we append one additional channel by replicating a pretrained channel kernel to initialize the NIR channel; for 2-pol SAR input (VV+VH), we retain the first two pretrained channel kernels to form $\mathbf{W}'\in\mathbb{R}^{C_{\mathrm{out}}\times 2\times k\times k}$ (for single-pol we only retain the first pretrained channel kernels). This preserves the pretrained weight distribution at the earliest layer while aligning the encoder with remote-sensing spectral/polarimetric inputs.

\textbf{Carrier-token hierarchical attention for efficient global semantics.}
At stage $l\in\{1,\dots,4\}$, each stream produces a multi-scale feature map $\mathbf{F}^{(l)}\in\mathbb{R}^{H_l\times W_l\times C_l}$, forming a four-stage pyramid. For notational convenience, we reshape the feature map into a token sequence
$\mathbf{X}^{(l)}=\mathrm{Tokenize}(\mathbf{F}^{(l)})\in\mathbb{R}^{N_l\times C_l}$,
where $N_l=H_lW_l$. Let $\mathbf{X}^{(l)}_{w_i}\in\mathbb{R}^{n_l\times C_l}$ denote the
subset of tokens within the $i$-th local window $w_i$ (with $n_l$ tokens per window).
Each feature map is partitioned into $M_l$ non-overlapping local windows $\{w_i\}_{i=1}^{M_l}$, and each window is associated with a compact carrier token $\mathbf{t}_i$ summarizing its semantics. The computation is decomposed into (i) local aggregation, (ii) global interaction on carrier tokens, and (iii) local injection:
\begin{equation}
\label{eq:local-global}
\left\{
\begin{aligned}
\mathbf{X}'^{(l)}_{w_i} &= \mathcal{F}_{\mathrm{local}}\!\left(\mathbf{X}^{(l)}_{w_i}; \mathbf{t}'_i\right)\\
\mathbf{t}_i &= \mathcal{A}_{\mathrm{local}}\!\left(\mathbf{X}^{(l)}_{w_i}\right)\\
\{\mathbf{t}'_i\}_{i=1}^{M_l} &= \mathcal{G}_{\mathrm{global}}\!\left(\{\mathbf{t}_j\}_{j=1}^{M_l}\right)
\end{aligned}
\right.,
\end{equation}
where $\mathbf{X}^{(l)}_{w_i}$ denotes the tokens within window $w_i$ at stage $l$. By confining dense interaction to local windows and performing global mixing on carrier tokens (rather than on all $N_l=H_lW_l$ tokens), HAT substantially reduces computation while retaining cross-window communication.

\textbf{Progressive tri-stream encoding.}
The Optical and SAR streams extract modality-specific pyramids $\{\mathbf{F}^{(l)}_{\mathrm{opt}}\}$ and $\{\mathbf{F}^{(l)}_{\mathrm{sar}}\}$, while the Cross-Modal stream maintains a progressively updated representation along the pyramid. Concretely, the stage-1 fused cross-modal feature is forwarded to the cross-modal encoder to generate deeper cross-modal features (e.g., at $\times 8, \times 16, \times 32$), which are further refined by subsequent stage-wise fusion. These multi-scale outputs serve as backbone features for the subsequent DCHF, which further suppresses cloud-contaminated optical responses during cross-modal interaction.

\subsection{Discrepancy-Conditioned Hybrid Fusion Mechanism}
Effective multimodal fusion critically depends on the quality of the global context descriptor used for channel recalibration. Conventional channel attention\cite{woo2018cbam,wang2020eca} typically employs Global Average Pooling (GAP) to aggregate spatial information into channel-wise statistics. 
However, under cloud occlusion where feature reliability becomes spatially non-uniform, GAP indiscriminately pools cloud-corrupted responses into the global statistics, biasing the resulting descriptors. This biased context then perturbs channel recalibration and can destabilize subsequent cross-modal interaction.

\begin{figure}[t]
  \centering
  \includegraphics[width=1\columnwidth]{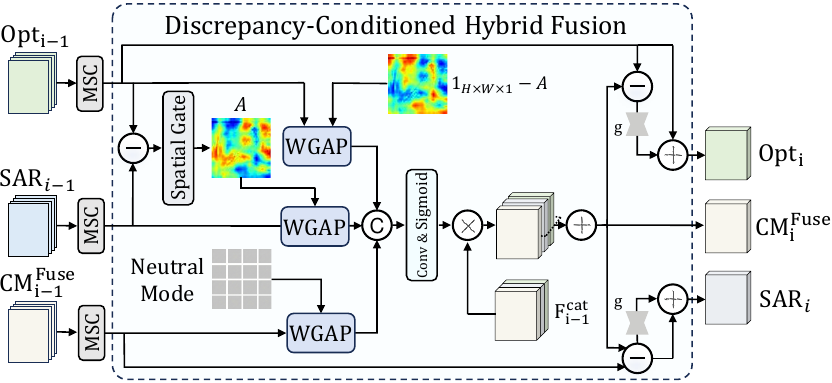}
  \caption{Architecture of the proposed DCHF module. It uses discrepancy-guided attention map $\mathbf{A}$ to perform weighted GAP for robust channel recalibration and fusion, producing $CM_i^{\text{Fuse}}$ and refined $\{Opt_i,SAR_i\}$ from $\{Opt_{i-1},SAR_{i-1},CM_{i-1}^{\text{Fuse}}\}$ (with $CM_{1}^{\text{Fuse}}$ empty). $\oplus/\ominus/\otimes$: element-wise ops; $\mathrm{C}$: concatenation.} 
  \label{fig:DCHF}
\end{figure}

To alleviate this issue, we propose a Discrepancy-Conditioned Hybrid Fusion (DCHF) module that incorporates an explicit \emph{cross-modal discrepancy attention} into global descriptor construction for robust channel recalibration. Let $\mathbf{F}_{\mathrm{opt}}, \mathbf{F}_{\mathrm{sar}}, \mathbf{F}_{\mathrm{fuse}}$ denote the feature tensors corresponding to $\{\mathrm{Opt}_{i-1}, \mathrm{SAR}_{i-1}, \mathrm{CM}_{i-1}^{\mathrm{Fuse}}\}$ in Fig.~\ref{fig:DCHF}.
Although the two modalities are encoded by parallel streams, SAR responses remain relatively stable under clouds, while optical features may deviate significantly due to cloud artifacts. Therefore, their discrepancy provides a practical cue to highlight regions where optical observations are less reliable and SAR cues are more informative. We compute a discrepancy map
\begin{equation}
\mathbf{D} = \psi(\mathbf{F}_{\mathrm{opt}}, \mathbf{F}_{\mathrm{sar}}),
\end{equation}
where $\psi(\cdot)$ is implemented as a lightweight difference operator $ \mathbf{F}_{\mathrm{opt}}-\mathbf{F}_{\mathrm{sar}}$.
This map is fed into a lightweight spatial gating network $f_{\text{sg}}(\cdot)$ (CBAM-style SpatialGate with $7\times 7$ convolutions) to produce a discrepancy attention map:
\begin{equation}
\mathbf{A} = \sigma\!\left(f_{\mathrm{sg}}(\mathbf{D})\right)\in[0,1]^{H\times W\times 1}.
\end{equation}
Here, $\mathbf{A}$ indicates stronger cross-modal inconsistency and thus a higher reliance on SAR cues, rather than a calibrated probabilistic uncertainty. For convenience, we further define an optical reliability map $\mathbf{R}_{\mathrm{opt}} = \mathbf{1} - \mathbf{A}$.

\textbf{Discrepancy-conditioned Global Descriptor (DCGD).}
Guided by $\mathbf{A}$, we replace standard GAP with a discrepancy-conditioned weighted GAP (WGAP) when constructing branch-wise channel descriptors. Specifically, we define
\begin{equation}
\mathbf{W}^{(\mathrm{opt})}_{\mathrm{rel}}=\mathbf{R}_{\mathrm{opt}},\quad
\mathbf{W}^{(\mathrm{sar})}_{\mathrm{rel}}=\mathbf{A},\quad
\mathbf{W}^{(\mathrm{cm})}_{\mathrm{rel}}=\mathbf{1}_{H\times W\times1},
\end{equation}
where $W_{\text{cm}}$ adopts a unit weight for the cross-modal stream. For each branch $b\in\{\text{opt},\text{sar},\text{cm}\}$, we compute
\begin{equation}
\hat z^{(b)}_c =
\frac{\sum_{i,j} \mathbf{W}^{(b)}_{\mathrm{rel}}(i,j)\, \mathbf{F}^{(b)}_{i,j,c}}
{\sum_{i,j} \mathbf{W}^{(b)}_{\mathrm{rel}}(i,j)+\epsilon_w},
\end{equation}
where $\mathbf{W}^{(b)}_{\mathrm{rel}}$ is selected according to $b\in\{\mathrm{opt},\mathrm{sar},\mathrm{cm}\}$,
and $\epsilon_w$ is a small constant. This asymmetric weighting prevents global descriptors from being dominated by cloud-corrupted optical regions: optical statistics are aggregated more conservatively from reliable (less inconsistent) regions, while SAR statistics emphasize regions where cross-modal inconsistency is high and SAR cues are most informative.

\textbf{Channel interaction and fusion injection.}
The descriptors from all branches are concatenated to form $\hat{\mathbf{z}}$, which is processed by an ECA-style 1D convolution\cite{wang2020eca} to capture local cross-channel dependencies. Here, $\mathrm{Conv1D}_k(\cdot)$ performs efficient local channel mixing on $\hat{\mathbf{z}}$, and $\sigma(\cdot)$ maps the response to a channel-wise gating vector $\mathbf{s}\in(0,1)^{|\hat{\mathbf{z}}|}$:
\begin{equation}
\mathbf{s} = \sigma\!\left(\mathrm{Conv1D}_k(\hat{\mathbf{z}})\right).
\end{equation}
The resulting channel-wise weights rescale the concatenated multimodal features (e.g., $\mathbf{F}_{\mathrm{opt}}$, $\mathbf{F}_{\mathrm{sar}}$, and $\mathbf{F}_{\mathrm{fuse}}$), producing a fused representation that is injected into the residual gating paths of the fusion block for progressive multi-scale fusion. In this way, DCHF couples discrepancy-conditioned descriptor construction with channel interaction, mitigating noise propagation under cloud occlusion.

\subsection{Dual-Task Decoding and Teacher-Guided Optimization}
\label{sec:method_D}

Given the discrepancy-conditioned fused representations, EDC adopts a lightweight U-Net decoder to recover spatial resolution. Through skip connections, the decoder combines high-level semantics from deep stages with low-level structural details from shallow stages, which helps preserve fine boundaries in dense land-cover mapping.
We attach two task-specific heads on top of the decoded features: (i) a semantic segmentation head that predicts land-cover logits, and (ii) an auxiliary cloud removal head that reconstructs the cloud-free optical image. This auxiliary reconstruction serves as an appearance-level regularizer, encouraging the encoder-decoder to preserve cloud-invariant surface cues that are beneficial for segmentation. For fair comparison with strong baselines, we keep the cloud removal and distillation strategy consistent with prior multi-task/distillation frameworks and focus our contributions on the efficiency-oriented encoder and the discrepancy-conditioned fusion mechanism.

\paragraph{Dual-task losses}
Let $y$ denote the ground-truth land-cover label map and $\hat{y}$ the predicted logits. We optimize the segmentation head using pixel-wise cross-entropy over valid pixels:
\begin{equation}
\mathcal{L}_{seg}=\frac{1}{|\Omega|}\sum_{(i,j)\in\Omega}\mathrm{CE}\!\left(\hat{y}_{i,j},y_{i,j}\right),
\end{equation}
where $\Omega$ is the valid pixel set (excluding void/unlabeled pixels).

For cloud removal, let $I^{cf}$ denote the ground-truth cloud-free optical image and $\hat{I}^{cr}$ denote the reconstructed output. We optimize the reconstruction branch using a cloud-mask-reweighted generalized Charbonnier loss\cite{barron2019robustloss} :
\begin{equation}
\label{eq:crloss}
\left\{
\begin{aligned}
\mathcal{L}_{cr} &= \frac{1}{CHW}\sum_{c,i,j}\left(1+\lambda M_{i,j}\right)\
\rho\!\left(\hat{I}^{cr}_{c,i,j}-I^{cf}_{c,i,j}\right)\\
\rho(d) &= (d^2+\epsilon^2)^{p}
\end{aligned}
\right.,
\end{equation}
where $M\in\{0,1\}^{H\times W}$ is the cloudmask ($M_{i,j}=1$ for cloudy pixels), and we set $\lambda=5$,$\quad p=0.45$, and $\epsilon=10^{-3}$. We upsample $\hat{I}^{cr}$ to match the ground-truth resolution before computing $\mathcal{L}_{cr}$ when needed.

\paragraph{Teacher-guided feature-level distillation}
Cloud occlusion can perturb both covered pixels and nearby feature interactions. To transfer clear-sky semantic priors, we employ a cloud-free teacher network trained on cloud-free inputs. Denote by $\Phi^S$ the student features and $\Phi^T$ the teacher features extracted from the corresponding layer (we distill the cross-modal decoded features). We align feature resolutions using a lightweight upsampling operator and impose a masked feature matching loss on clear pixels:
\begin{equation}
\label{eq:kdloss}
\left\{
\begin{aligned}
\mathcal{L}_{kd} &= \frac{1}{|\Omega_{kd}|}\sum_{(i,j)\in\Omega_{kd}}
\left\|\Phi^S_{i,j}-\Phi^T_{i,j}\right\|_2^2\\
\Omega_{kd} &= \{(i,j)\mid M_{i,j}=0\}
\end{aligned}
\right..
\end{equation}
We apply the distillation loss only on clear pixels, because these regions provide reliable teacher-student semantic correspondence. Enforcing strict feature matching in cloud corrupted regions may introduce undesirable bias due to the severe input discrepancy between the cloudy student and the cloud-free teacher.

\paragraph{Overall objective}
The final training objective is
\begin{equation}
\mathcal{L}_{total}=\mathcal{L}_{seg}+\beta\,\mathcal{L}_{cr}+\gamma\,\mathcal{L}_{kd},
\end{equation}
where $\beta$ and $\gamma$ balance the auxiliary reconstruction and distillation terms.

\section{EXPERIMENTS}
\subsection{Experimental settings}
\subsubsection{Datasets}
We conduct experiments on two datasets:

\medskip
{\bf{M3M-CR}}\cite{xu2024aligncr} is built from 780 disjoint regions of interest (ROIs) sampled worldwide. The original benchmark focuses on cloud removal and provides, for each ROI, a quartet of orthorectified, geo-referenced images: a cloudy optical image, a cloud-free optical image, a Sentinel-1 SAR image, and a land-cover map derived from ESA WorldCover. In our study, these scenes are tiled into 63{,}000 patches. 60{,}000 patches from 660 ROIs are used for model training, while 3{,}000 patches from the remaining 120 ROIs form the held-out test set. Within the training set, we further hold out 6{,}494 patches from 65 ROIs for validation, and use the remaining 53{,}506 patches from 595 ROIs for training. Each patch contains a $300 \times 300$ PlanetScope optical tile with four spectral bands (blue, green, red, and near-infrared) at $3\,\mathrm{m}$ ground sampling distance, and a co-registered Sentinel-1 GRD SAR tile acquired in IW mode with VV and VH polarizations at $10\,\mathrm{m}$ resolution. The SAR images are originally $90 \times 90$ pixels. We upsampled them by nearest-neighbor interpolation to match the $300 \times 300$ optical grid.

\medskip
{\bf{WHU-OPT-SAR}}\cite{li2022_mcanet} is a large-scale optical-SAR land-use dataset collected over Hubei Province, China. It comprises 100 co-registered GF-1 optical images and GF-3 SAR images, each of size $5{,}556 \times 3{,}704$ pixels and covering in total roughly $50{,}000~\text{km}^2$ with $5\,\mathrm{m}$ spatial resolution after resampling. The optical data provide four channels (RGB and near-infrared), while the SAR data is single polarization. Pixel-level land-use annotations with multiple categories are available for all scenes. For our experiments, we first align the optical and SAR images at $5\,\mathrm{m}$ resolution, then uniformly crop them into non-overlapping $256 \times 256$ patches. We keep all four optical bands together with the SAR image as network inputs, and randomly split the patches into training, validation, and test subsets at the scene level to avoid spatial leakage between splits. Following CloudSeg, we synthesized cloudy conditions by uniformly adding Perlin noise to each spectral band of the optical images. The original cloud-free optical patches were retained as supervision targets and as references for evaluation under ideal conditions.

\subsubsection{Evaluation Metrics}
To evaluate \textbf{EDC} and enable fair comparisons with existing methods, we adopt commonly used metrics for both tasks in our dual-task setting, i.e., \textit{land-cover semantic segmentation} and \textit{cloud removal}.

\textbf{Evaluation Metrics for Land-Cover Mapping.}
We employ two standard segmentation metrics, Mean Intersection over Union (mIoU) and Mean Pixel Accuracy (mPA), to evaluate region overlap and class-wise pixel accuracy, respectively. Let $C$ denote the number of land-cover classes, and $\mathrm{TP}_i$, $\mathrm{FP}_i$, and $\mathrm{FN}_i$ denote the numbers of true positive, false positive, and false negative pixels for class $i$, respectively. 

mIoU is the primary indicator for semantic segmentation, measuring the spatial overlap between the predicted map and the ground truth. It explicitly penalizes both false positives and false negatives by taking the intersection-to-union ratio for each class and then averaging over all classes. The calculation is as follows: \begin{equation} \mathrm{mIoU}=\frac{1}{C}\sum_{c=1}^{C}\frac{\mathrm{TP}_c}{\mathrm{TP}_c+\mathrm{FP}_c+\mathrm{FN}_c}. \end{equation} 

And mPA complements mIoU by evaluating the proportion of correctly classified pixels within each ground-truth class and then averaging across classes, thus treating all categories equally regardless of class frequency. The calculation is as follows: \begin{equation} \mathrm{mPA}=\frac{1}{C}\sum_{c=1}^{C}\frac{\mathrm{TP}_c}{\mathrm{TP}_c+\mathrm{FN}_c}. \end{equation} Higher mIoU and mPA indicate better land-cover mapping performance.

\textbf{Evaluation Metrics for Cloud Removal.}
For cloud removal, we evaluate the fidelity of the reconstructed cloud-free optical output $\hat{\mathbf{Y}}\in\mathbb{R}^{C\times H\times W}$ against the cloud-free reference $\mathbf{Y}\in\mathbb{R}^{C\times H\times W}$ using three widely adopted image-quality metrics: Peak signal-to-noise ratio (PSNR), Structural Similarity Index (SSIM) and Mean Absolute Error (MAE). Let $\mathrm{MAX}$ denote the maximum valid pixel value.

PSNR measures the reconstruction fidelity by quantifying the ratio between the maximum possible signal power and the reconstruction noise power, where a larger PSNR indicates less distortion. We first compute the mean squared error (MSE) between $\hat{Y}$ and ${Y}$, and then derive PSNR accordingly. The calculation procedure is shown as below:
\begin{equation}
\label{eq:mse-psnr}
\left\{
\begin{aligned}
\mathrm{PSNR} &= 10\log_{10}\left(\frac{\mathrm{MAX}^2}{\mathrm{MSE}}\right)\\\mathrm{MSE} &= \frac{1}{CHW}\sum_{c=1}^{C}\sum_{i=1}^{H}\sum_{j=1}^{W}
\left(\hat{Y}_{c,i,j}-Y_{c,i,j}\right)^2
\end{aligned}
\right..
\end{equation}



\begin{table*}[t]
\centering
\caption{Segmentation performance comparison under cloudy, cloud-free, and overall splits on M3M-CR and WHU-OPT-SAR between EDC and state-of-the-art methods. Best results are in bold and the second-best are underlined.}
\label{tab:segmentation_mpa_miou}
\small
\setlength{\tabcolsep}{4pt}
\renewcommand{\arraystretch}{1.15}

\begin{tabular}{lcccccccccccc}
\toprule
\multirow{3}{*}{\textbf{Method}} &
\multicolumn{6}{c}{\textbf{M3M-CR}} &
\multicolumn{6}{c}{\textbf{WHU-OPT-SAR}} \\
\cmidrule(lr){2-7}\cmidrule(lr){8-13}
& \multicolumn{2}{c}{Cloudy} & \multicolumn{2}{c}{Cloud-Free} & \multicolumn{2}{c}{Overall}
& \multicolumn{2}{c}{Cloudy} & \multicolumn{2}{c}{Cloud-Free} & \multicolumn{2}{c}{Overall} \\
\cmidrule(lr){2-3}\cmidrule(lr){4-5}\cmidrule(lr){6-7}
\cmidrule(lr){8-9}\cmidrule(lr){10-11}\cmidrule(lr){12-13}
& mPA & mIoU & mPA & mIoU & mPA & mIoU & mPA & mIoU & mPA & mIoU & mPA & mIoU \\
\midrule
DCSA-Net\cite{rs14194941}     & 50.35 & 39.52 & 51.21 & 40.21 & 51.04 & 40.09 & 46.67 & 36.76 & 47.88 & 38.48 & 47.28 & 37.63 \\
AMM-FuseNet\cite{rs14184458}    &59.17&46.18 &63.74 &49.95  &61.77&48.29&51.68&32.21&55.83&38.19&53.97&35.24 \\
MCANet\cite{li2022_mcanet}      &61.17&49.16&64.75&51.52&63.29&50.62&41.04&31.39&41.34&31.92&41.19&31.66  \\
FTransUNet\cite{ma2024ftransunet}&60.57&48.89&66.49&54.60&63.37&50.63&56.12&45.10&58.79&47.43&57.47&46.28\\
CMX\cite{zhang2023cmxcrossmodalfusionrgbx}  &61.08&48.73&65.33&51.98&63.46&50.58&50.48&40.09&53.86&42.07&52.16&41.15\\
CloudSeg\cite{xu2024_cloudseg}    & \underline{63.43} & \underline{50.87} & \underline{69.02} & \underline{55.91} & \underline{66.57} & \underline{53.64} &
                   \underline{55.79} &\underline{45.01}&63.00&\underline{52.18}& 59.69& \underline{48.70}\\
\midrule

EDC*      & 62.16 & 49.63 & 67.94 & 54.43 & 65.38 & 52.29 & 
55.34 & 44.27 & \underline{64.30} & 51.88 & \underline{59.87} & 48.11 \\

\textbf{EDC (Ours)} &
\textbf{64.95} & \textbf{51.24} & \textbf{71.05} & \textbf{56.69} & \textbf{68.33} & \textbf{54.20} & \textbf{57.24} & \textbf{45.74} & \textbf{66.01} & \textbf{53.40} & \textbf{61.67} & \textbf{49.58}\\
\bottomrule
\end{tabular}
\end{table*}


SSIM measures image quality by comparing the similarity in three major aspects: luminance, contrast, and structure, which are more consistent with human visual perception than pixel-wise errors. The calculation procedure is shown as below:
\begin{equation}
\mathrm{SSIM}(\hat{\mathbf{Y}},\mathbf{Y})=
\frac{(2\mu_{\hat{Y}}\mu_{Y}+c_1)(2\sigma_{\hat{Y}Y}+c_2)}
{(\mu_{\hat{Y}}^2+\mu_{Y}^2+c_1)(\sigma_{\hat{Y}}^2+\sigma_{Y}^2+c_2)},
\end{equation}
where $\mu$ and $\sigma^2$ denote the mean and variance, $\sigma_{\hat{Y}Y}$ is the covariance, and $c_1,c_2$ are small constants for numerical stability. For multi-spectral images, SSIM is computed per band and then averaged over $C$ bands.


MAE measures the average absolute reconstruction deviation between $\hat{\mathbf{Y}}$ and $\mathbf{Y}$, providing an intuitive pixel-wise error magnitude that is easy to interpret; a smaller MAE indicates higher reconstruction accuracy. The calculation procedure is shown as below:
\begin{equation}
\mathrm{MAE}=\frac{1}{CHW}\sum_{c=1}^{C}\sum_{i=1}^{H}\sum_{j=1}^{W}\left|\hat{Y}_{c,i,j}-Y_{c,i,j}\right|.
\end{equation}
Higher PSNR/SSIM and lower MAE indicate better cloud removal performance.

\subsubsection{Implementation Details} Our EDC is trained and evaluated using an NVIDIA A100 80GB in the PyTorch framework. We trained the network using AdamW with a weight decay of $1\times10^{-4}$. The learning rate was set to $3\times10^{-4}$. The batch size was 160 and the maximum number of training epochs was set to 100. The weighting factors $\beta$ and $\gamma$ were empirically set to 1.

\subsection{Comparisons with semantic segmentation methods}
We compare EDC with representative multimodal semantic segmentation baselines, including DCSA-Net\cite{rs14194941}, MCANet\cite{li2022_mcanet}, AMM-FuseNet\cite{rs14184458}, FTransUNet\cite{ma2024ftransunet}, CMX\cite{zhang2023cmxcrossmodalfusionrgbx}, and CloudSeg
\cite{xu2024_cloudseg}. Following the common evaluation protocol in cloud-robust land-cover mapping, we additionally report results on cloudy and cloud-free subsets determined by the provided cloud coverage maps, besides the overall performance. Quantitative comparisons on M3M-CR are summarized in Tab.~\ref{tab:segmentation_mpa_miou}. For clarity, we also include a simplified variant, EDC*, which performs semantic segmentation only, without the auxiliary cloud removal branch or teacher-guided distillation used in the full model.

As shown in Tab.~\ref{tab:segmentation_mpa_miou}, EDC achieves the best overall segmentation performance, reaching 68.33 mPA and 54.20 mIoU. More importantly, its advantage remains consistent across different cloud conditions. In cloudy regions, EDC outperforms CloudSeg and all other baselines, indicating stronger robustness when optical observations are severely degraded. In cloud-free regions, it still delivers the best accuracy, showing that the proposed design improves cloud resilience without compromising clear-sky performance.

These improvements are consistent with the proposed efficiency-oriented context aggregation and discrepancy-conditioned fusion. By conditioning global descriptor construction and channel recalibration on a cross-modal discrepancy attention, EDC reduces the influence of cloud-induced outliers and selectively strengthens complementary SAR structural cues in inconsistent regions, leading to more stable predictions in cloudy areas. Moreover, the hierarchical context modeling of the encoder enables occluded locations to borrow semantic priors from distant clear-sky patches, alleviating local ambiguity. Finally, the performance gap between EDC and EDC* further highlights the benefit of the multi-task and teacher-guided training strategy with consistent improvements in both cloudy and cloud-free subsets.

\begin{figure*}[!t]
  \centering
  \includegraphics[width=0.98\linewidth]{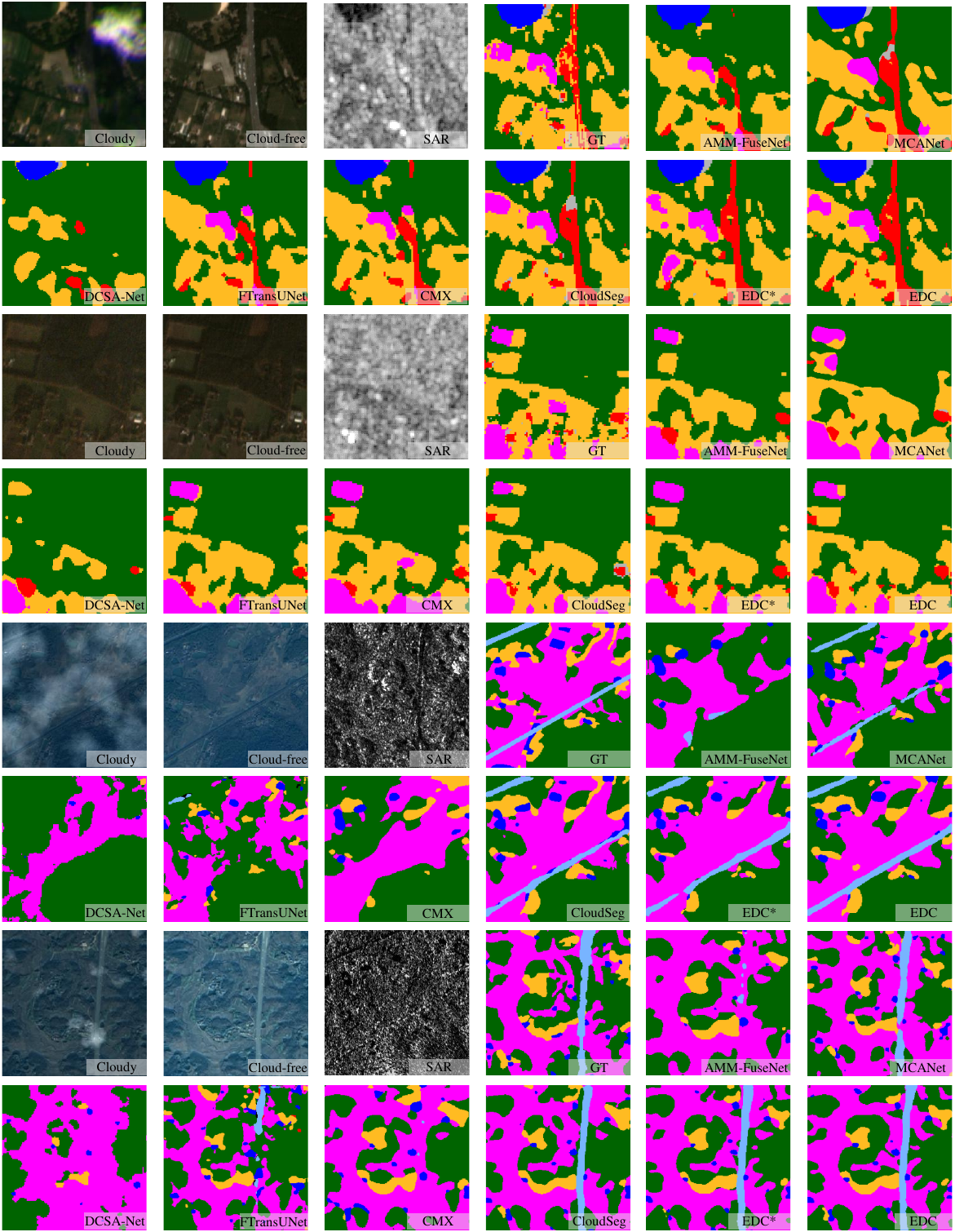}
  \caption{Visualization of land cover mapping for 4 different scenes. The first two scenes are from the M3M-CR dataset, while the latter two are from the WHU-OPT-SAR dataset. For each scene, from top-left to bottom-right are respectively the cloudy image, the cloud-free image, the SAR image, the ground truth, the result from AMM-FuseNet, MCANet, DCSA-Net, FTransUNet, CMX, CloudSeg and our EDC.}
  \label{fig:vis}
\end{figure*}



\begin{table*}[t]
\centering
\caption{Quantitative comparisons between the proposed EDC and the current state-of-the-art methods in the field of cloud removal. The results with the highest performance are shown in bold, while the second-best results are underlined.}
\label{tab:segcr_quantitative_comparison}
\renewcommand{\arraystretch}{1.15}
\setlength{\tabcolsep}{9pt}

\begin{tabular}{lccc ccc}
\toprule
\multirow{2}{*}{Method} &
\multicolumn{3}{c}{M3M-CR} &
\multicolumn{3}{c}{WHU-OPT-SAR} \\
\cmidrule(lr){2-4}\cmidrule(lr){5-7}
& SSIM$\uparrow$ & PSNR$\uparrow$ (\text{dB}) &
MAE$\downarrow$ ($10^{-2}$)
& SSIM$\uparrow$ & PSNR$\uparrow$ ($dB$) &
MAE$\downarrow$ ($10^{-2}$)
\\
\midrule

HS2P\cite{li2023hs2p}        & 0.8809 & 29.61& 2.72 & 0.8093&28.73&2.96 \\
USSRN-CR\cite{wang2024ussrncr}& 0.8942 & 29.99 & 2.64 &0.8122&28.94&2.87  \\
CloudSeg\cite{xu2024_cloudseg}    & 0.8907 & 29.56 & 2.69 & 0.8167 & 28.61 & 2.89 \\
Align-CR\cite{xu2024aligncr}    & 0.9050 & 29.12 & 2.57 &0.8192  &29.42&2.65  \\
GLF-CR\cite{xu2022glfcr} & \underline{0.9076} &\underline{30.25} & \underline{2.43} & \underline{0.8215}&\underline{29.92}&\underline{2.56}\\
\midrule
\textbf{EDC (Ours)}
                & \textbf{0.9185} & \textbf{30.94} & \textbf{2.26}
                & \textbf{0.8236} & \textbf{30.01} & \textbf{2.49} \\
\bottomrule
\end{tabular}
\end{table*}


Fig.~\ref{fig:vis} presents qualitative comparisons on four representative scenes from M3M-CR and WHU-OPT-SAR. Overall, EDC produces predictions that are more consistent with the ground truth than the competing methods, especially under severe cloud contamination. Although EDC* yields broadly similar boundary layouts, it remains less accurate in semantic discrimination within cloud-affected regions. In the first M3M-CR scene, EDC more completely recovers spatially coherent road structures, even when the roads are heavily obscured in the cloudy optical image but remain visible in SAR. In the second M3M-CR scene, EDC also more clearly separates rangeland from agricultural land, whereas the other methods, including EDC*, exhibit boundary confusion and incomplete partitioning. In the first WHU-OPT-SAR scene, where near-complete cloud cover severely degrades optical cues, EDC produces more continuous roads and cleaner water bodies. This improvement is consistent with the role of DCHF, which suppresses cloud-corrupted optical responses and relies more on stable SAR cues in the affected regions.

Beyond segmentation, we also evaluate the auxiliary cloud removal branch of EDC against representative reconstruction baselines, including HS2P\cite{li2023hs2p}, USSRN-CR\cite{wang2024ussrncr}, CloudSeg\cite{xu2024_cloudseg}, Align-CR\cite{xu2024aligncr}, and GLF-CR\cite{xu2022glfcr}. Tab.~\ref{tab:segcr_quantitative_comparison} shows that EDC achieves the best reconstruction quality on both datasets. In particular, it consistently surpasses the strongest competitor, GLF-CR, on M3M-CR, and maintains the top ranking on WHU-OPT-SAR. This result indicates that the auxiliary branch serves not only as a training regularizer, but also as an effective cloud removal module for recovering high-fidelity cloud-free imagery.

\subsection{Ablation Study}
\label{sec:ablation}
We systematically evaluate the proposed method from three perspectives: 1) the overall advantage of the full architecture over a naive baseline; 2) the fine-grained effectiveness of the discrepancy-conditioned fusion design, including DCHF and DCGD; and 3) the contribution of the dual-task optimization strategy, including knowledge distillation and auxiliary cloud removal. All quantitative results on M3M-CR and WHU-OPT-SAR are summarized in Tab.~\ref{tab:ablation}.

\begin{table}[t]
\centering
\caption{Ablation study of different architectural components and training strategies on M3M-CR and WHU-OPT-SAR datasets.}
\label{tab:ablation}
\renewcommand{\arraystretch}{1.15}
\setlength{\tabcolsep}{10pt}

\begin{tabular}{lcccc}
\toprule
\multirow{2}{*}{Method} & \multicolumn{2}{c}{M3M-CR} & \multicolumn{2}{c}{WHU-OPT-SAR} \\
\cmidrule(lr){2-3}\cmidrule(lr){4-5}
& mPA & mIoU & mPA & mIoU \\
\midrule
w/o \textit{DCHF}          & 66.84 & 53.35 & 60.06&48.91  \\
w/o \textit{DCGD}          & 67.09 & 53.44 &60.57&48.99  \\
w/o \textit{Distillation}  & 65.43 & 52.54 &61.00&48.90  \\
w/o \textit{Cloud Removal} & 65.97 & 52.85 &61.21& 49.26\\
\textit{EDC-naive}& 64.88 & 52.01 & 59.37&47.29\\
\midrule
\textbf{EDC (Ours)}      & \textbf{68.33} & \textbf{54.20} & \textbf{61.67}&\textbf{49.58}\\
\bottomrule
\end{tabular}
\end{table}

\paragraph{Effectiveness of the Proposed Architecture}
To verify the advantage of the full framework, we compare EDC with a baseline variant, EDC-naive. EDC-naive removes the auxiliary supervision, including teacher-guided distillation and cloud removal, and replaces the proposed discrepancy-conditioned fusion with standard Squeeze-and-Excitation (SE) blocks\cite{hu2018senet}. It can therefore be regarded as a further degraded version of EDC*. 
As reported in Tab.~\ref{tab:ablation}, the complete model yields clear and consistent gains over EDC-naive on both datasets. On M3M-CR, EDC improves mIoU by 2.19\% and mPA by 3.45\%, with similar improvements also observed on WHU-OPT-SAR. These results suggest that combining discrepancy-conditioned fusion with auxiliary supervision is beneficial under the spatially non-stationary corruption introduced by clouds. Without explicit reliability-aware correction, EDC-naive is more easily affected by cloud-dominated features and therefore tends to accumulate errors in occluded regions. By contrast, DCHF suppresses discrepancy-induced optical outliers during fusion, while the auxiliary cloud removal task and teacher-guided distillation further regularize cloud-invariant features and reliable clear-sky semantics, leading to more consistent representations and more robust final predictions.

\begin{figure*}[t]
  \centering
  \includegraphics[width=0.9\linewidth]{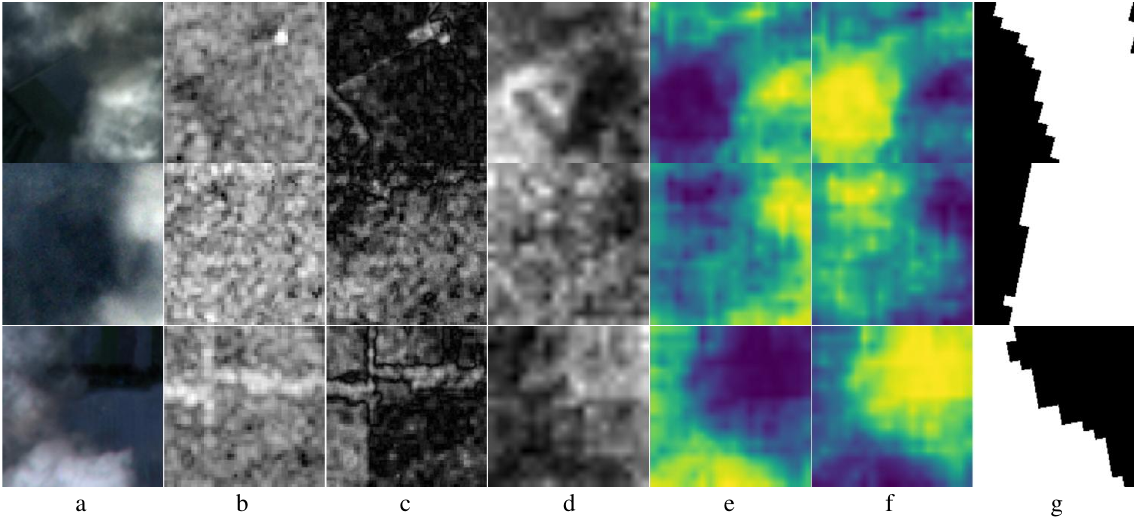}
  \caption{Intermediate outputs of the DCHF module are shown from left to right as follows: (a) cloudy optical image, (b) SAR image, (c) discrepancy map between cloud-free optical image and SAR image, (d) discrepancy map between cloudy optical image and SAR image, (e) discrepancy attention map, (f) reliability map, (g) cloudmask image.}
  \label{fig:midresult}
\end{figure*}

\paragraph{Detailed Contributions of DCHF and DCGD}
To further analyze the proposed fusion design, we separately evaluate the contributions of the DCHF module and its core component, DCGD.

\noindent\textbf{The Effectiveness of DCHF:}
We replace DCHF with a standard SE-based fusion block to evaluate the role of discrepancy-guided fusion. Unlike DCHF, the SE block does not perform discrepancy-conditioned spatial filtering. As shown in Tab.~\ref{tab:ablation}, this substitution leads to a clear performance drop, with the mIoU on M3M-CR decreasing from 54.20\% to 53.35\%. This result supports the value of spatial modulation guided by the cross-modal discrepancy map. Visual evidence in Fig.~\ref{fig:midresult} is consistent with this observation: compared with the intrinsic modality difference shown for reference in Fig.~\ref{fig:midresult}(c), the discrepancy map under cloudy observations in Fig.~\ref{fig:midresult}(d) exhibits pronounced responses concentrated in cloud-covered regions. Accordingly, the spatial gating network produces an attention map in Fig.~\ref{fig:midresult}(e) that highlights corrupted regions and suppresses the propagation of cloud-induced noise during spatial feature interaction.



\noindent\textbf{The Effectiveness of DCGD:}
To assess the role of discrepancy-conditioned global descriptor construction, we replace DCGD with standard GAP while keeping the remaining fusion design unchanged. Under this setting, the mIoU on M3M-CR decreases to 53.44\%, indicating that standard GAP is more easily affected by cloud artifacts that are indiscriminately pooled into global channel statistics and can bias subsequent channel recalibration. This interpretation is further supported by the optical reliability map in Fig.~\ref{fig:midresult}(f) ($1-\mathbf{A}$), where low-reliability regions align well with the ground-truth cloudmask in Fig.~\ref{fig:midresult}(g). This observation suggests that DCGD helps suppress cloud-corrupted optical contributions during global descriptor construction. By dynamically down-weighting inconsistent optical regions, DCGD refines the global descriptors and encourages channel recalibration to rely more on stable SAR structural cues, thereby helping preserve semantic representations under severe occlusion.


\paragraph{Effectiveness of Knowledge Distillation and Multi-Task Regularization}
Finally, we analyze the contribution of the teacher-student framework and the auxiliary task to the final land-cover mapping performance.

\noindent\textbf{Knowledge Distillation:} Removing the feature-level distillation branch leads to a clear performance drop on both benchmarks. As shown in Tab.~\ref{tab:ablation}, the mPA on M3M-CR decreases from 68.33\% to 65.43\%. On WHU-OPT-SAR, mPA and mIoU decrease by 0.67\% and 0.68\%, respectively. These results suggest that clear-pixel distillation from the cloud-free teacher helps stabilize semantic representation learning by providing reliable semantic anchors in unoccluded regions. This supervision indirectly improves the final predictions under occlusion.

\noindent\textbf{Cloud Removal:}
The exclusion of the auxiliary cloud removal head also adversely affects the segmentation accuracy. Specifically, the mIoU on M3M-CR decreases from 54.20\% to 52.85\%, and the mIoU on WHU-OPT-SAR also drops. This validates the role of the cloud removal head as an effective appearance-level regularizer. By forcing the encoder to reconstruct the underlying cloud-free optical image, the auxiliary task encourages the model to capture fine-grained surface details and cloud-invariant structural cues. This multi-task optimization strategy creates a synergy where the low-level geometric information recovered by the CR branch directly facilitates the high-level semantic discrimination required for robust land-cover mapping under cloudy conditions.

\subsection{Efficiency Analysis and Architectural Pareto Optimality}
\label{sec:efficiency_pareto}
To assess the deployment practicality of EDC, we compare model complexity (Params/GMAC), inference throughput (Thrp.), and segmentation accuracy (mIoU) on M3M-CR and WHU-OPT-SAR. All throughput results are measured under a fixed protocol with $160\times160$ inputs, batch size 128, and FP16 inference with AMP. The quantitative results in Tab.~\ref{tab:efficiency_miou} summarize the practical accuracy--efficiency trade-off among current cloud-resilient optical-SAR segmentation models.

\subsubsection{Pareto Frontier under a Fixed Deployment Protocol}
Under the same deployment protocol, EDC achieves the best segmentation accuracy on both datasets while also remaining the fastest among all compared methods. This places EDC on a favorable Pareto frontier: unlike heavy context-aggregation models, it does not trade runtime efficiency for stronger accuracy, and unlike lightweight alternatives, it does not sacrifice segmentation quality for speed. These results indicate that EDC achieves a more effective balance between accuracy and deployment efficiency than existing baselines.



\medskip

\begin{table}[t]
\centering
\caption{Model complexity, computation, inference throughput, and segmentation accuracy comparison on M3M-CR and WHU-OPT-SAR. Thrp.\ denotes inference throughput in images per second (img/s), measured under the benchmarking protocol in Sec.~\ref{sec:efficiency_pareto}.}

\label{tab:efficiency_miou}
\scriptsize
\renewcommand{\arraystretch}{1.10}
\setlength{\tabcolsep}{4pt}

\begin{tabular}{lccccc}
\toprule
\multirow{2}{*}{Method} & \multirow{2}{*}{Para (M)} & \multirow{2}{*}{GMAC} & \multirow{2}{*}{Thrp.} & \multicolumn{1}{c}{M3M-CR} & \multicolumn{1}{c}{WHU-OPT-SAR} \\
\cmidrule(lr){5-5}\cmidrule(lr){6-6}
& & & & mIoU & mIoU \\
\midrule
DCSA-Net        & 80.01  & 8.17  & 275.30 & 40.09 & 37.63 \\
AMM-FuseNet     & \textbf{62.30} & 43.21 & 464.3 & 48.29 & 35.24 \\
MCANet          & 131.76 & 11.90 & \underline{952.3}& 50.62 & 31.66 \\
FTransUNet      & 203.40 & 21.90 & 410.39 & 50.63 & 46.28 \\
CMX             & 66.57  & \textbf{5.58} &968.3& 50.58 & 41.15 \\
CloudSeg        & 203.21 & 16.25 & 491.2 & 53.64 & 48.70 \\
\midrule
\textbf{EDC (Ours)} & 108.36 & 10.74 &\textbf{971.2}& \textbf{54.20} & \textbf{49.58} \\
\bottomrule
\end{tabular}
\end{table}

\subsubsection{Efficiency from Structural Decoupling in the Encoder}
The efficiency advantage mainly comes from the proposed \emph{Efficiency-Oriented Multi-Scale Encoder} (Sec.~III--B). Instead of relying on dense global token interaction as in heavy Transformer-style backbones, EDC routes long-range context through \emph{Carrier Tokens}, which decouples local feature extraction from global dependency modeling. This design reduces redundant global mixing and leads to a better complexity--throughput profile in practice. Compared with CloudSeg, EDC uses substantially fewer parameters and less computation, while still achieving higher mIoU on both M3M-CR and WHU-OPT-SAR. The corresponding throughput gain further confirms that the architectural efficiency translates into consistent runtime benefits under the same inference setting.

\subsubsection{Effective Capacity under Cloud Occlusion via DCHF}
Beyond raw complexity, robust performance under cloud occlusion also depends on how effectively computation is utilized. Methods that aggregate context uniformly over all spatial locations may not only incur unnecessary overhead, but also propagate cloud-corrupted optical responses into the fused representation. This tendency can be observed in methods with relatively high computational cost but limited accuracy gains. In contrast, EDC employs discrepancy-conditioned fusion (Sec.~III--C) to suppress unreliable optical features and rely more on stable SAR cues in occluded regions. This improves the effective use of its parameter and computation budget, enabling stronger segmentation performance without requiring a heavier model.

\subsubsection{Budgeted Performance: Lightweight vs.\ Heavy vs.\ Ours}
Tab.~\ref{tab:efficiency_miou} and Fig.~\ref{fig:pareto} also reveal a clear two-sided limitation in existing designs. Lightweight models can be fast, but often lack sufficient representational capacity for cloud-occluded scene understanding, leading to inferior segmentation accuracy. Conversely, heavy baselines allocate substantially larger parameter and computation budgets, yet still fail to outperform EDC while running much more slowly. EDC provides a better balance between these two extremes: it retains sufficient capacity for long-range context reasoning and reliable fusion under occlusion, while structural optimization avoids the redundancy of heavier architectures, yielding stronger accuracy at practical deployment speed.

\begin{table*}[t]
\centering
\caption{Representation efficiency diagnostics. We report effective-rank statistics and activation sparsity at the deepest cross-modal encoder stage (\texttt{stage4\_cm}, $C{=}512$), and the channel dimension $C_d$ and linear CKA with logits at the cross-modal decoder output (\texttt{cm\_decoder\_out}).}
\label{tab:repr_diag}
\renewcommand{\arraystretch}{1.15}
\setlength{\tabcolsep}{6pt}
\footnotesize
\begin{tabular}{lccc|cc}
\toprule
\multirow{2}{*}{Method} & \multicolumn{3}{c|}{\texttt{stage4\_cm}} & \multicolumn{2}{c}{\texttt{cm\_decoder\_out}} \\
\cmidrule(lr){2-4} \cmidrule(lr){5-6}
 & erank $\uparrow$ & nerank $\uparrow$ & sparsity ($|x|<10^{-3}$) & $C_d$ & CKA $\uparrow$ \\
\midrule
CMX & 8.70 & 0.017 & 0.004 & 512 & 0.609 \\
CloudSeg & 23.58 & 0.046 & 0.667 & 16 & 0.821 \\
EDC (Ours) & \textbf{131.97} & \textbf{0.258} & 0.119 & 32 & \textbf{0.930} \\
\bottomrule
\end{tabular}
\end{table*}

\subsection{Representation-Level Efficiency Analysis}
To complement the complexity and mIoU comparisons, we further examine how EDC organizes multimodal representations relative to representative baselines. We extract features from (i) the deepest cross-modal fusion stage (\texttt{stage4\_cm}, $C{=}512$) and (ii) the cross-modal decoder output (\texttt{cm\_decoder\_out}). For a fair layer-wise comparison, we use stage-4 features for all methods, since both the SegFormer-based baselines (CMX and CloudSeg) and our FasterViT-based encoder follow a four-stage hierarchy. Therefore, \texttt{stage4\_cm} in EDC is aligned with the stage-4 fused representations in the baselines. We then compute the \textit{Effective Rank} (erank) and \textit{Normalized Effective Rank} (nerank) to measure representational diversity and channel utilization, and use linear Centered Kernel Alignment (CKA) between \texttt{cm\_decoder\_out} and logits to assess task alignment\cite{kornblith2019cka}.
Let \(X \in \mathbb{R}^{C\times N}\) denote a feature matrix with \(C\) channels and \(N\) spatial locations where the spatial dimensions are flattened. We first compute the covariance

\begin{equation}
    \Sigma = \frac{1}{N} X X^\top,
\end{equation}
with eigenvalues \(\{\lambda_i\}_{i=1}^C\), and define the normalized spectrum as,
\begin{equation}
\label{eq:pi-entropy}
\left\{
\begin{aligned}
H(p) &= -\sum_{i=1}^C p_i \log p_i\\
p_i &= \frac{\lambda_i}{\sum_{j=1}^C \lambda_j}
\end{aligned}
\right..
\end{equation}
The effective rank and normalized effective rank are then
\begin{equation}
\label{eq:erank-nerank}
\left\{
\begin{aligned}
\operatorname{nerank}(X) &= \frac{\operatorname{erank}(X)}{C}\\
\operatorname{erank}(X) &= \exp\!\big(H(p)\big)
\end{aligned}
\right..
\end{equation}
Given two centered feature matrices \(X \in \mathbb{R}^{N\times d_x}\) and \(Y \in \mathbb{R}^{N\times d_y}\) collected over the same set of pixels, the linear CKA is defined as
\begin{equation}
\operatorname{CKA}(X,Y) = 
\frac{\big\|X^\top Y\big\|_F^2}
     {\big\|X^\top X\big\|_F \,\big\|Y^\top Y\big\|_F }.
\end{equation}
In this subsection, \(X\) denotes the layer features and \(Y\) denotes the logits.

\textbf{Preventing Feature Degeneration in Deep Layers:}
As shown in Tab.~\ref{tab:repr_diag}, CloudSeg exhibits reduced representational diversity at the deepest encoder stage. Although its fused feature has 512 channels, the normalized effective rank is only 0.046, together with high activation sparsity (66.7\%), suggesting that many channel directions are weakly utilized during deep fusion. In contrast, EDC maintains substantially higher-rank fused features (erank $=131.97$, nerank $=0.258$) with much lower sparsity ($11.9\%$), indicating better preservation of diverse feature directions in deep cross-modal fusion.

\textbf{Structural Efficiency and Bottleneck Alignment:}
The comparison at the decoder output reveals distinct structural behaviors. CMX maintains a high-dimensional pre-logit feature ($C_d{=}512$) but achieves a relatively low CKA (0.609), indicating weaker linear alignment at this stage. CloudSeg employs a much narrower bottleneck ($C_d{=}16$) and is associated with higher alignment (0.821) compared to CMX. Notably, EDC compresses the deep fused representation into a compact pre-logit bottleneck ($C_d{=}32$) while achieving the highest task alignment (0.930). Taken together, these results indicate that EDC not only mitigates deep representational degeneration in the fused encoder, but also converts the fused representation into a compact and highly task-aligned pre-logit space, consistent with a more efficient (closer-to-linear) readout in the final segmentation head.

Overall, EDC maintains higher-rank and less sparse deep fusion features, and achieves the strongest logit alignment with a compact bottleneck ($C_d{=}32$), supporting a more efficient organization of multimodal representations under a constrained complexity budget.

\subsection{Confidence and Calibration Analysis}
\label{sec:calibration}

Beyond standard accuracy metrics like mIoU, the reliability of predictions is also paramount. A reliable model should be well-calibrated, meaning its predicted confidence probabilities should match its actual accuracy\cite{niculescu2005probabilities}. To evaluate this, we compute pixel-wise confidence as the maximum softmax probability and group predictions into confidence bins\cite{hendrycks2017baseline}. For each bin, we report the empirical accuracy (fraction of correctly classified pixels) and the average confidence. We visualize the calibration behavior using residual calibration plots, defined as \( \mathrm{Residual} = \mathrm{Accuracy} - \mathrm{Confidence}. \)
Graphically, a perfectly calibrated model yields a flat line at zero; values below zero indicate over-confidence, while values above zero indicate under-confidence. As illustrated in Tab.~\ref{tab:ece_strong}, we also report the Expected Calibration Error (ECE) to quantify the overall deviation\cite{guo2017calibration,pavlovic2025understandingmodelcalibration}.
The analysis is performed on three subsets of the test set: all pixels, only cloud-free pixels, and only cloudy pixels, according to the ground-truth cloudmask.
To compute ECE, we partition the confidence interval \([0,1]\) into \(B\) bins. Let \(\mathcal{S}_b\) be the set of pixels whose confidence falls into bin \(b\), with \(n_b=|\mathcal{S}_b|\) and \(N=\sum_{b=1}^B n_b\). Denoting the average confidence and empirical accuracy in bin \(b\) as \(\operatorname{conf}_b\) and \(\operatorname{acc}_b\), the ECE is
\begin{equation}
\operatorname{ECE}=\sum_{b=1}^{B}\frac{n_b}{N}\,\big|\operatorname{acc}_b-\operatorname{conf}_b\big|.   
\end{equation}

\begin{figure}[!t]
\centering
\includegraphics[width=\columnwidth]{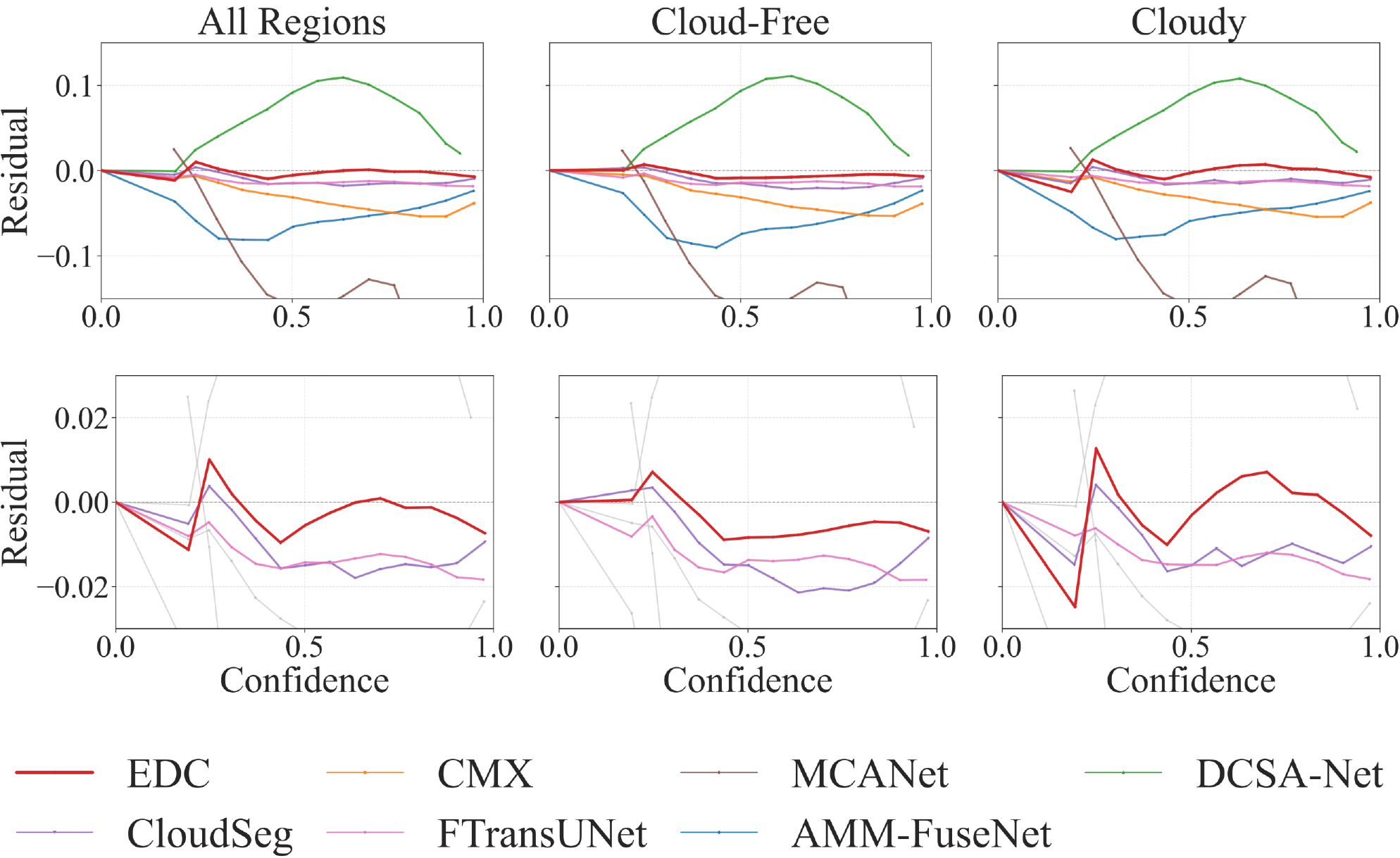}
\caption{Residual calibration results on the M3M-CR dataset. The top row shows residual calibration plots of all compared methods, while the bottom row zooms in on the top-performing models for fine-grained comparison. From left to right are the All Regions, Cloud-Free, and Cloudy subsets, where negative indicates over-confidence and positive indicates under-confidence.}
\label{fig:calib}
\end{figure}

\subsubsection{Overall Calibration Landscape}
Fig.~\ref{fig:calib} (top) visualizes the calibration behavior of all compared methods. The results reveal distinct behavioral tiers. Earlier fusion baselines exhibit pronounced miscalibration: DCSA-Net is predominantly under-confident, yielding a relatively high ECE of 0.0745, whereas MCANet tends to be over-confident and frequently assigns high confidence to incorrect predictions. In contrast, more recent methods, including CloudSeg, FTransUNet, and EDC, remain much closer to the zero-residual line, indicating substantially improved calibration quality.

\subsubsection{Robustness Under Cloud Occlusion}
To examine the differences among the strongest models more closely, Fig.~\ref{fig:calib} (bottom) zooms in on the residual calibration curves of EDC, CloudSeg, and FTransUNet, with particular attention to the challenging Cloudy subset. Both CloudSeg and FTransUNet exhibit consistently negative residuals in cloudy regions, indicating a tendency toward over-confidence under occlusion. This effect is especially evident for CloudSeg, whose curve falls noticeably below zero and whose Cloudy ECE reaches 0.0120. Such behavior suggests that, without explicit discrepancy modeling, dense context aggregation may over-smooth corrupted features and produce high-confidence errors in cloud-affected regions\cite{ovadia2019uncertaintyshift}.

In contrast, EDC shows the most stable calibration behavior, with residuals remaining close to zero across confidence intervals. Notably, in the high-confidence range (\(>0.8\)) of the Cloudy subset, EDC avoids the pronounced negative deviation observed in the baselines. This visual trend is consistent with the quantitative results: EDC achieves the lowest Cloudy ECE of \textbf{0.0053}, corresponding to a \textbf{55.8\% reduction} relative to CloudSeg.

These observations are consistent with the design goal of discrepancy-conditioned fusion in reducing over-confident errors under cloud occlusion. By using the cross-modal discrepancy map as a spatial gate, EDC suppresses cloud-corrupted optical responses before fusion, which helps produce confidence estimates that more closely track empirical accuracy in occluded regions.

\begin{table}[t]
\centering
\caption{ECE of strong baselines on M3M-CR. ALL/CF/CY denote the all/cloud-free/cloudy regions.}
\label{tab:ece_strong}
\setlength{\tabcolsep}{8pt}
\renewcommand{\arraystretch}{1.15}
\footnotesize
\begin{tabular}{lccc}
\toprule
Method & ECE (All) $\downarrow$ & ECE (CF) $\downarrow$ & ECE (CY) $\downarrow$ \\
\midrule
EDC (Ours)     & \textbf{0.0046} & \textbf{0.0065} & \textbf{0.0053} \\
CloudSeg    & 0.0126          & 0.0133          & 0.0120          \\
FTransUNet  & 0.0161          & 0.0163          & 0.0159          \\
CMX         & 0.0420          & 0.0420          & 0.0419          \\
\bottomrule
\end{tabular}
\end{table}

\section{Conclusion}
\label{sec:calibration}

In this paper, we presented EDC, an efficiency-oriented, discrepancy-conditioned framework for cloud-resilient optical-SAR semantic segmentation via progressive spatial-spectral fusion. It is built on two core designs: (i) EOME introduces Carrier Tokens to decouple local feature extraction from global dependency modeling, alleviating the efficiency-receptive-field bottleneck; and (ii) DCHF uses pixel-wise cross-modal discrepancies as a reliability proxy to gate cloud-corrupted responses and prevent noise from biasing global context aggregation.

Experiments on M3M-CR and WHU-OPT-SAR demonstrate state-of-the-art segmentation accuracy, both in cloud-covered and cloud-free regions. Meanwhile, EDC remains highly efficient, reducing parameters by 46.7\% and improving inference speed by 97.7\% compared with heavy context-aggregation models such as CloudSeg, indicating that strong cloud-resilient mapping does not require prohibitive computational cost.

Looking forward, we will optimize EDC for resource-constrained on-board satellite processing to enable real-time cloud-resilient monitoring at the sensor edge, and further explore lightweight compression techniques to facilitate this deployment.

\bibliographystyle{IEEEtran}
\bibliography{refs}

\end{document}